\documentclass[conference]{IEEEtran}

\usepackage[sort,compress]{cite}
\usepackage[pdftex]{graphicx}
\usepackage{amsmath,amssymb}
\usepackage{algorithmic}
\usepackage{array}
\usepackage{url}
\usepackage{color,xspace}
\usepackage{adjustbox}
\usepackage{multirow}
\usepackage{tabularx}
\usepackage{booktabs}
\usepackage{listings}
\usepackage{algorithm,algorithmic}
\usepackage[keeplastbox]{flushend}

\newcommand{\vct}[1]{\ensuremath{\boldsymbol{#1}}} %for greek letters

\newcommand{\set}[1]{\ensuremath{\mathcal{#1}}}
\newcommand{\con}[1]{\ensuremath{\mathsf{#1}}}
\newcommand{\T}{\ensuremath{\top}}

\newcommand{\myparagraph}[1]{\smallskip \noindent \textbf{#1.}}
\newcommand{\ie}{{i.e.}\xspace}
\newcommand{\eg}{{e.g.}\xspace}

\newcommand{\aka}{{a.k.a.}\xspace}

\newcommand{\Drebin}{\textrm{Drebin}\xspace}

\newcommand{\DDrebin}{\textsl{Drebin}\xspace}

\newcommand{\SVM}{{SVM}\xspace}
\newcommand{\SVMRBF}{{SVM-RBF}\xspace}
\newcommand{\RF}{{RF}\xspace}

\newcommand{\manifest}{\texttt{manifest}\xspace}
\newcommand{\dexcode}{\texttt{dexcode}\xspace}

\begin{document}

\title{Explaining Black-box Android Malware Detection}

\author{
\IEEEauthorblockN{
Marco Melis\IEEEauthorrefmark{1},
Davide Maiorca\IEEEauthorrefmark{1},
Battista Biggio\IEEEauthorrefmark{1}\IEEEauthorrefmark{2},
Giorgio Giacinto\IEEEauthorrefmark{1}\IEEEauthorrefmark{2} and
Fabio Roli\IEEEauthorrefmark{1}\IEEEauthorrefmark{2}}
\IEEEauthorblockA{\IEEEauthorrefmark{1}DIEE, University of Cagliari, Piazza d'Armi, 09123, Cagliari\\ \{marco.melis,davide.maiorca,battista.biggio,giacinto,roli\}@diee.unica.it}
\IEEEauthorrefmark{2} Pluribus One, Italy}

\maketitle

\begin{abstract} 
Machine-learning models have been recently used for detecting malicious Android applications, reporting impressive performances on benchmark datasets, even when trained only on features statically extracted from the application, such as system calls and permissions.
However, recent findings have highlighted the fragility of
such in-vitro evaluations with benchmark datasets, showing that very few changes to the
content of Android malware may suffice to evade detection.
How can we thus trust that a malware detector performing well on benchmark data will continue to do so when deployed in an operating environment?
To mitigate this issue, the most popular Android malware detectors use linear, explainable machine-learning models to easily identify the most influential features contributing to each decision.
In this work, we generalize this approach to any black-box machine-learning model, by leveraging a gradient-based approach to identify the most influential local features. This enables using nonlinear models to potentially increase accuracy without sacrificing interpretability of decisions.
Our approach also highlights the global characteristics learned by the model to discriminate between benign and malware applications. Finally, as shown by our empirical analysis on a popular Android malware detection task, it also helps identifying
potential vulnerabilities of linear and nonlinear models against adversarial manipulations.
\end{abstract}

\IEEEpeerreviewmaketitle

\section{Introduction}

With more than $400$ millions of malicious applications discovered in the wild, Android malware constitutes one of the major threats in mobile security. Among the various detection strategies proposed by companies and academic researchers, those based on machine learning have shown the most promising results, due to their flexibility against malware variants and obfuscation attempts~\cite{rieck14-drebin,chen16-asiaccs}. 
Despite the impressive performances reported by such approaches on benchmark datasets, the problem of Android malware detection in the wild is still far from being solved. 
The validity of such optimistic, in-vitro evaluations has been indeed questioned from recent adversarial analyses showing that only few changes to the content of a malicious Android application may suffice to evade detection by a learning-based detector~\cite{demontis17-tdsc,calleja18}.
Besides this fragility to well-crafted evasion attacks (\aka adversarial examples)~\cite{biggio18,biggio13-ecml,szegedy14-iclr,goodfellow15-iclr}, Sommer and Paxson~\cite{sommer10} have more generally questioned the suitability of black-box machine-learning approaches to computer security. In particular, how can we thus trust the predictions of a machine-learning model \emph{in vivo}, \ie, when it is deployed in an operating environment, to take subsequent reliable actions? How can we understand whether we are selecting a proper model before deployment? How about its security properties against adversarial attacks?

To partially address these issues, Android malware detectors often restrict themselves to the use of \emph{linear}, \emph{explainable} machine-learning models that allow one to easily identify the most influential features contributing to each decision (Sect.~\ref{sect:drebin})~\cite{rieck14-drebin,backes17}.
More generally, \emph{intepretability} of machine-learning models has recently become a relevant research direction to more thoroughly address and mitigate the aforementioned issues, especially in the case of \emph{nonlinear} \emph{black-box} machine-learning algorithms~\cite{lipton16,baehrens10-jmlr,koh17-icml,ribeiro16,kim17-arxiv}. 
Some approaches aim to explain \emph{local} predictions (\ie, on each specific sample) by identifying the most influential features~\cite{baehrens10-jmlr,ribeiro16} or prototypes from training data~\cite{koh17-icml}. 
Others have proposed techniques and methodologies towards providing \emph{global explanations} about the salient characteristics learned by a given machine-learning algorithm~\cite{lipton16,kim17-arxiv}.

In this work, we generalize current explainable Android malware detection approaches to \emph{any} black-box machine-learning model, by leveraging a gradient-based approach to identify the most influential \emph{local} features (Sect.~\ref{sect:interpreting}).
For non-differentiable learning algorithms, like decision trees, we extract gradient information by learning a differentiable approximation. Notably, this idea has originally been exploited to construct gradient-based evasion attacks against non-differentiable learners, and evaluate their \emph{transferability}, \ie, the probability that an attack crafted against a learning algorithm succeeds against a different one~\cite{biggio13-ecml,goodfellow15-iclr,russu16-aisec}.
Accordingly, our approach provides interpretable decisions even for Android malware detectors exploiting nonlinear learning algorithms to potentially increase detection accuracy. 
Moreover, by averaging the local relevant features across different classes of samples, our approach allows also highlighting the \emph{global} characteristics learned by a given model to identify benign applications and different classes of Android malware.

We perform our experimental analysis with a popular Android malware detector named \Drebin~\cite{rieck14-drebin} (Sect.~\ref{sect:exp}). It extracts information the Android application through static analysis, 
and provides interpretable decisions by leveraging a linear classification algorithm.
To test the validity of our approach, we show how to retain the interpretability of \Drebin on nonlinear algorithms, including Support Vector Machines (SVMs) and Random Forests (RFs).
Interestingly, we also show that the interpretations provided by our approach can help identifying potential vulnerabilities of both linear and nonlinear Android malware detectors against adversarial manipulations.

We conclude the paper by discussing contributions and limitations of this work, and future research directions towards developing more robust malware detectors (Sect.~\ref{sect:conclusions}).

\section{Android Malware Detection}
\label{sect:drebin}

In this section, we provide some background on how Android applications are structured, and
then discuss \Drebin~\cite{rieck14-drebin}, the malware detector used in our analysis.

\subsection{Android Background}
\label{sect:android_malware}

Android applications are \texttt{apk} files, \ie, zipped archives that must contain two files: the Android \manifest and the classes.dex. Additional \texttt{xml} and resource files are respectively used to define the application layout and to provide multimedia contents. As \Drebin only analyzes the Android \manifest and classes.dex files, we briefly describe them below.

\myparagraph{Android Manifest} The \manifest file holds information about how the application is organized in terms of its \emph{components}, \ie, parts of code that perform specific actions; \eg, one component might be associated to a screen visualized by the user (\emph{activity}) or to the execution of audio in the background (\emph{services}). It is also possible to perform actions on the occurrence of a specific event (\emph{receivers}). The actions of each component are further specified through \emph{filtered intents}; \eg, when a component sends data to other applications, or is invoked by a browser. Special types of intent filters (\eg, LAUNCHER) can specify that a certain component is executed as soon as the application is opened. %Special types of components are \emph{entry points}, \ie, activities, services and receivers that are loaded when requested by a specific filtered intent (\eg, an activity is loaded when an application is launched, and a service is activated when the device is turned on).
The \manifest also contains the list of \emph{hardware components} and \emph{permissions} requested by the application to work (\eg, Internet access).

\myparagraph{Dalvik Bytecode (\dexcode)} The classes.dex file
embeds the compiled source code of an application, including all the user-implemented methods and classes.  %Such file is often  converted to .jar for further analyses by using tools such as \texttt{dex2jar}.\footnote{\url{https://github.com/pxb1988/dex2jar}}
%For the purposes of our paper, some elements of the file are
%particularly of interest, as they are often exploited by
%malicious applications.
Classes.dex may contain specific API
calls that can access sensitive resources such as personal
contacts (\emph{suspicious calls}). Additionally, it contains all system-related,
\emph{restricted API calls} whose functionality require
\emph{permissions} (\eg, using the Internet). Finally, this
file can contain references to \emph{network addresses} that
might be contacted by the application.

\subsection{Drebin}

\begin{figure*}[t]
	\begin{center}
		\includegraphics[width=0.9\textwidth]{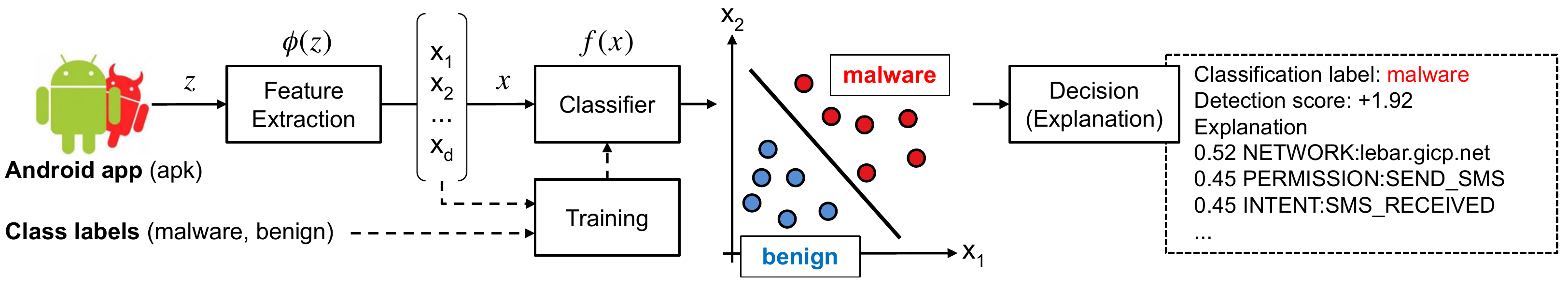}
		\caption{A schematic representation of \Drebin, adapted from~\cite{demontis17-tdsc}. First, applications are represented as binary vectors in a $\con d$-dimensional feature space. A linear classifier is then trained on an available set of malware and benign applications, assigning a weight to each feature. During classification, unseen applications are scored by the classifier by summing up the weights of the present features: if $f(\vct x) \geq 0$, they are classified as malware. \Drebin also explain each decision by reporting the most suspicious (or benign) features present in the app, along with the weight assigned to them by the linear classifier~\cite{rieck14-drebin}.}
		\vspace{-18pt}
		\label{fig:system-arch}
	\end{center}
\end{figure*}

\Drebin performs a lightweight static analysis of Android applications.
The extracted features are used to embed benign and malware apps into a high-dimensional
vector space, train a machine-learning model, and then perform classification of never-before-seen apps.
An overview of the system architecture is given in Fig.~\ref{fig:system-arch}, and discussed more in detail below.

\myparagraph{Feature extraction} First, \Drebin statically analyzes a set of available
Android applications to construct a suitable feature space.
All features extracted by \Drebin are presented as \emph{strings} and
organized in 8 different feature sets, as listed in Table \ref{tab:feature_sets}.
\begin{table}[b]
	\caption{Overview of feature sets.}
	\centering
	\begin{tabular}{ ll|ll }
		\toprule
		\multicolumn{2}{ c| }{\texttt{manifest}} & \multicolumn{2}{ c }{\texttt{dexcode}} \\
		\midrule
		$S_{1}$ & Hardware components & $S_{5}$ & Restricted API calls\\
		$S_{2}$ & Requested permissions & $S_{6}$ & Used permission \\
		$S_{3}$ & Application components & $S_{7}$ & Suspicious API calls \\
		$S_{4}$ & Filtered intents & $S_{8}$ & Network addresses\\
		\bottomrule
	\end{tabular}
	\label{tab:feature_sets}
\end{table}
Android applications are then mapped onto the feature space as follows.
Let us assume that an app is
represented as an object $\vct z \in \set Z$, being $\set Z$
the abstract space of all \texttt{apk} files.
We then denote with $\Phi : \set Z \mapsto \set X$ a function
that maps an \texttt{apk} file $\vct z$ to a $\con d$-dimensional feature vector $\vct
x = ( x^{1}, \ldots, x^{\con d} )^{\T} \in \set X=\{0,1\}^{\con d}$,
where each feature is set to 1 (0) if the corresponding \emph{string} is present (absent) in
the \texttt{apk} file $\vct z$.
An application encoded in feature space may thus look like the following:\vspace{-1em}
\begin{center}
\resizebox{\linewidth}{!}{
  \begin{minipage}{\linewidth}
\begin{align}
\nonumber
\small \vct x = \Phi( \vct z) \mapsto
\begin{pmatrix}
\cdots \\ \small 0\\ \small 1\\
\cdots \\ \small 1\\ \small 0\\
\cdots\\
\end{pmatrix}
\begin{array}{ll}
\cdots & \multirow{4}{*}{\hspace{-1mm}\bigg \} $S_2$ }\\
\texttt{\small permission::SEND\_SMS} \\
\texttt{\small permission::READ\_SMS}\\
\cdots & \multirow{4}{*}{\hspace{-1mm}\bigg \} $S_5$ }\\
\texttt{\small api\_call::getDeviceId}\\
\texttt{\small api\_call::getSubscriberId}\\
\cdots & \\
\end{array}
\end{align}
  \end{minipage}
}\end{center}
\vspace{1em}

\myparagraph{Learning and Classification}
\Drebin uses a linear SVM to perform detection. 
It can be expressed in terms of
a linear function $f : \set X \mapsto \mathbb R$, \ie, 
$f(\vct x) = \vct w^{\T}\vct x + b$, where  $\vct w \in \mathbb R^{\con d}$ denotes the vector of \emph{feature weights},
and $b \in \mathbb R$ is the so-called \emph{bias}.
These parameters, optimized during training, identify a hyperplane that separates the two classes in feature space.
During classification, unseen apps are then classified as malware if $f(\vct x) \geq 0$, and as benign otherwise. 

\myparagraph{Explanation} \Drebin explains its decisions by reporting, for any given application, the most influential features, \ie, the features that are present in the given application and are assigned the highest absolute weights by the classifier.
 For instance, in Fig.~\ref{fig:system-arch}, it is easy to see, from its most influential features,
 that a malware sample is correctly identified by \Drebin as it connects to a suspicious URL and uses SMS as a side channel for communication.
As we aim to extend this approach to nonlinear models, in this work we also consider an
SVM with the Radial Basis Function (RBF) kernel and a random forest to learn nonlinear functions $f(\vct x)$.

\section{Interpreting Decisions of Learning-based Black-box Android Malware Detectors}
\label{sect:interpreting}

We discuss here our idea to generalize the explainable decisions of Drebin and other locally-explainable Android malware detectors~\cite{rieck14-drebin,backes17} to any black-box (\ie, nonlinear) machine-learning algorithm.
In addition, we also propose a method to explain the \emph{global} characteristics influencing the decisions of the learning-based malware detector at hand.

\begin{figure}[t]
\centering
\includegraphics[width=0.35\textwidth]{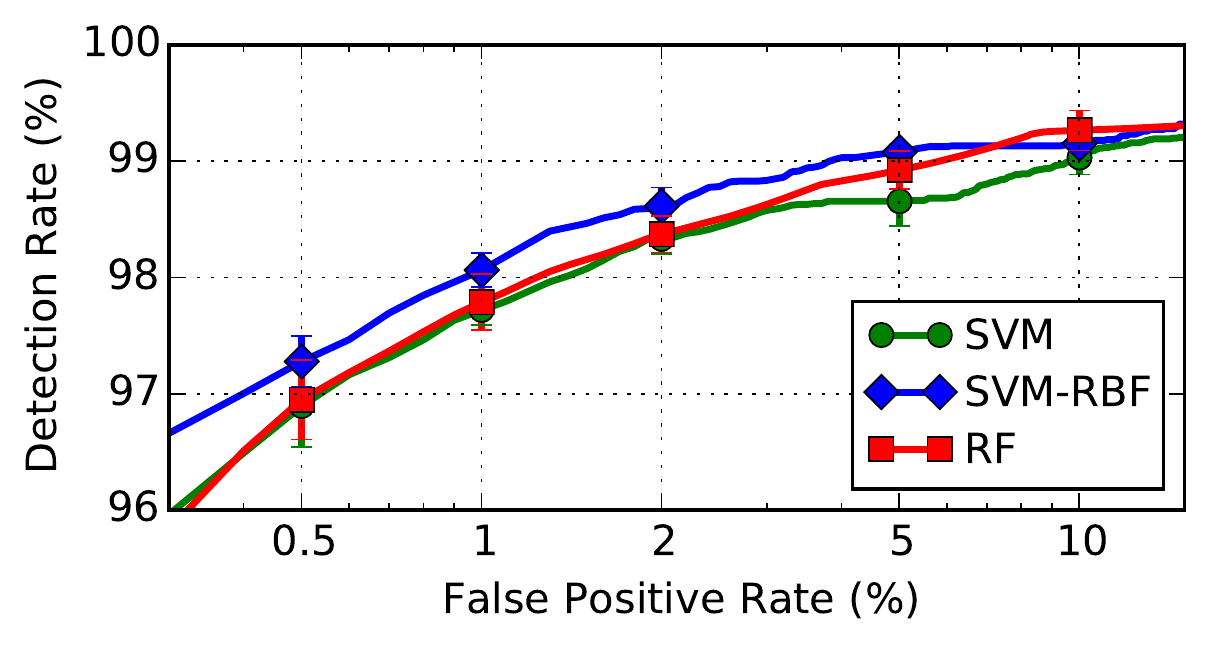}
\vspace{-5pt}
\caption{Average ROC curves for the given classifiers on the \DDrebin data.}\label{fig:roc-mean}
\vspace{-2pt}
\end{figure}

\myparagraph{Local explanations} Previous work has highlighted that gradients and, more generally, linear approximations computed around the input point $\vct x$ convey useful information for explaining the local predictions provided by a learning algorithm~\cite{baehrens10-jmlr,ribeiro16}.
The underlying idea is to identify as \emph{most influential} those features associated to the highest (absolute) values of the local gradient $\nabla f(\vct x)$, being $f$ the confidence associated to the predicted class.
However, in the case of sparse data, as for Android malware, these approaches tend to identify a high number of influential features which are \emph{not} present in the given application, thus making the corresponding predictions difficult to interpret.
For this reason, in this work we consider a slightly different approach, inspired from the notion of directional derivative.
In particular, we project the gradient $\nabla f(\vct x)$ onto $\vct x$ to obtain a \emph{feature-relevance} vector $\vct \nu = \nabla f(\vct x) \cdot \vct x \in \mathbb R^{\con d}$, where $\cdot$ denotes the element-wise product. We then normalize $\vct \nu$ to have a unary $\ell_1$ norm, \ie, $\vct r = \vct \nu / \|  \vct \nu \|_1$, to ensure that only non-null features in $\vct x$ are identified as relevant for the decision. Finally, the absolute values of $\vct r$ can be ranked in descending order to identify the most influential \emph{local} features.

\myparagraph{Global explanations} In contrast to other locally-explainable malware detectors~\cite{rieck14-drebin,backes17}, we also provide a \emph{global} analysis of the interpretability of the considered machine-learning models, aimed to identify the most influential features, on average, which characterize benign and malware samples.
Our idea is simply to average the relevance vectors $\vct r$ over different samples, \eg, separately for benign and malware data. Then, as in the local case, the absolute values of the average relevance vector $\bar{\vct r}$ can be ranked in descending order to identify the most influential \emph{global} features.

\myparagraph{Non-differentiable models} Our approach works under the assumption that $f(\vct x)$ is \emph{differentiable} and that its gradient $\nabla f(\vct x)$ is sufficiently smooth to provide meaningful information at each point.
When $f(\vct x)$ is not differentiable (\eg, for decision trees and random forests), or its gradient vanishes (\eg, if $f(\vct x)$ becomes constant in large regions of the input space), we compute approximate feature-relevance vectors by means of surrogate models. The idea is to train a differentiable approximation $\hat f(\vct x)$ of the target function $f(\vct x)$, similar to what has been done in~\cite{baehrens10-jmlr} for interpretability of non-differentiable models, and in~\cite{biggio13-ecml,russu16-aisec} to craft gradient-based evasion attacks against non-differentiable learning algorithms.
For instance, to reliably estimate a non-differentiable algorithm $f(\vct x)$ (\eg, a random forest), one can train a nonlinear SVM on a training set relabeled with the predictions provided by $f(\vct x)$~\cite{russu16-aisec}. 

\section{Experimental Analysis} \label{sect:exp}

In this section, we use our approach to provide local and global explanations for linear and nonlinear (including non-differentiable) classifiers trained on the features used by \Drebin. As we will see, this will also reveal some insights on their security against adversarial manipulations~\cite{demontis17-tdsc,calleja18}.

\myparagraph{Datasets} We use here the \DDrebin data~\cite{rieck14-drebin}, consisting of $121,329$ benign applications and $5,615$ malicious samples, labeled with VirusTotal. A sample is labeled as malicious if it is detected by at least five anti-virus scanners, whereas it is labeled as benign if no scanner flagged it as malware.

\myparagraph{Training-test splits} We average our results on 5 runs. In each run, we randomly select 60,000 apps from the \DDrebin data to train the learning algorithms, and use the rest for testing.

\myparagraph{Classifiers} We compare the standard \DDrebin implementation based on a linear SVM (\SVM) against an SVM with the RBF kernel (\SVMRBF) and a (non-differentiable) Random Forest (\RF). As discussed in Sect.~\ref{sect:interpreting}, a surrogate model is needed to interpret the \RF; to this end, we train an SVM with the RBF kernel on the training set relabeled by the \RF (yielding an approximation with accuracy higher than $99\%$ on average on the relabeled testing sets). The Receiver Operating Characteristic (ROC) curve for each classifier, averaged over the 5 repetitions, is reported in Fig. \ref{fig:roc-mean}.

\myparagraph{Parameter setting} We optimize the parameters of each classifier through a 3-fold cross-validation procedure.
In particular, we optimize $C \in \{10^{-2}, 10^{-1}, \ldots, 10^{2}\}$ for both linear and non-linear SVMs,
the kernel parameter $\gamma \in \{10^{-4}, 10^{-3}, \ldots, 10^{2}\}$ for the \SVMRBF,
and the number of estimators $n \in \{5,10, \ldots, 30\}$ for the \RF.

\begin{table*}[t]
\centering
\caption{Top-10 influential features for \SVM (top row) and \RF (bottom row) on ($i$) a benign sample (first column), ($ii$) a malware sample of the \texttt{SmsWatcher} family (second column), and ($iii$) a malware sample of the \texttt{Plankton} family (third column). The probability of each feature being present in bening ($p_B$) and malware ($p_M$) is also reported.}
\label{tab:local-ranks}
\begin{adjustbox}{width=0.32\textwidth}
\begin{tabular}{cp{6cm}rrr}
\toprule
   \textbf{Set} & \textbf{Feature Name}                                                                        & \multicolumn{1}{c}{$\vct r$ (\%)} & \multicolumn{1}{c}{$p_B$ (\%)} & \multicolumn{1}{c}{$p_M$ (\%)} \\
\midrule
     S2 & SEND\_SMS                     &  26.89 &     3.19 &    53.89 \\
     S4 & LAUNCHER                 & -15.40 &    96.42 &    93.56 \\
     S6 & SEND\_SMS                                       &   9.42 &     3.11 &    44.76 \\
     S2 & GET\_ACCOUNTS                 &   8.61 &     2.57 &     8.06 \\
     S8 & code.google.com                                            &   6.38 &     1.10 &     1.73 \\
     S7 & Ljava/io/IOException;-\ensuremath{>}printStackTrace        &   6.09 &    49.82 &    66.85 \\
     S2 & READ\_CONTACTS                &  -4.61 &     7.25 &    23.75 \\
     S2 & INTERNET                     &   4.30 &    83.29 &    96.26 \\
     S8 & ajax.googleapis.com                                        &  -3.19 &     0.76 &     0.54 \\
     S4 & android.intent.action.MAIN                       &   2.91 &    97.52 &    95.88 \\
\bottomrule
\end{tabular}
\end{adjustbox}
\hspace{0.05em}
\begin{adjustbox}{width=0.32\textwidth}
\begin{tabular}{cp{6cm}rrr}
\toprule
   \textbf{Set} & \textbf{Feature Name}                                                                        & \multicolumn{1}{c}{$\vct r$ (\%)} & \multicolumn{1}{c}{$p_B$ (\%)} & \multicolumn{1}{c}{$p_M$ (\%)} \\
  \midrule
     S2 & SEND\_SMS                       & 10.94 &     3.19 &    53.89 \\
     S3 & com.rjblackbox.swl.SMSActivity                         &  9.72 &     0.00 &     0.04 \\
     S3 & com.rjblackbox.swl.SMSForwarder                         &  9.72 &     0.00 &     0.04 \\
     S3 & com.rjblackbox.swl.SettingsActivity                    &  9.72 &     0.00 &     0.04 \\
     S4 & android.provider.Telephony.SMS\_RECEIVED            &  8.13 &     1.09 &    20.08 \\
     S4 & LAUNCHER                   & -6.26 &    96.42 &    93.56 \\
     S2 & RECEIVE\_SMS                    & -4.82 &     2.43 &    38.36 \\
     S6 & SEND\_SMS                                         &  3.83 &     3.11 &    44.76 \\
     S7 & Lorg/apache/http/client/methods/HttpPost         &  3.52 &    29.95 &    51.89 \\
     S7 & android/telephony/SmsMessage;-\ensuremath{>}createFromPdu & -3.51 &     1.44 &    16.19 \\
\bottomrule
\end{tabular}
\end{adjustbox}
\hspace{0.05em}
\begin{adjustbox}{width=0.32\textwidth}
\begin{tabular}{cp{6cm}rrr}
\toprule
   \textbf{Set} & \textbf{Feature Name}                                                                        & \multicolumn{1}{c}{$\vct r$ (\%)} & \multicolumn{1}{c}{$p_B$ (\%)} & \multicolumn{1}{c}{$p_M$ (\%)} \\
  \midrule
     S7 & TelephonyManager;-\ensuremath{>}getNetworkOperator     &  3.00 &     6.01 &    46.57 \\
     S4 & LAUNCHER                              & -2.50 &    96.42 &    93.56 \\
     S7 & TelephonyManager;-\ensuremath{>}getNetworkOperatorName & -2.46 &     5.08 &    28.99 \\
     S6 & ACCESS\_NETWORK\_STATE                                        & -2.32 &    47.92 &    56.40 \\
     S7 & android/net/Uri;-\ensuremath{>}fromFile                                  &  2.13 &    16.81 &    43.10 \\
     S2 & INSTALL\_SHORTCUT (launcher)   &  2.04 &     1.51 &    26.37 \\
     S2 & READ\_HISTORY\_BOOKMARKS (browser)        &  1.73 &     0.52 &    17.89 \\
     S5 & LocationManager;-\ensuremath{>}isProviderEnabled               & -1.70 &    12.53 &    17.12 \\
     S7 & com.apperhand.device.android.AndroidSDKProvider                     &  1.70 &     0.00 &    10.95 \\
     S7 & java/lang/reflect/Method;-\ensuremath{>}getReturnType                    & -1.52 &     5.97 &    12.22 \\
\bottomrule
\end{tabular}
\end{adjustbox}
\\\vspace{0.2em}\hspace{0pt}
\begin{adjustbox}{width=0.32\textwidth}
\begin{tabular}{cp{6cm}rrr}
\toprule
   \textbf{Set} & \textbf{Feature Name} & \multicolumn{1}{c}{$\vct r$ (\%)} & \multicolumn{1}{c}{$p_B$ (\%)} & \multicolumn{1}{c}{$p_M$ (\%)} \\
\midrule
     S2 & SEND\_SMS                                        &  25.82 &     3.19 &    53.89 \\
     S4 & LAUNCHER                                    & -18.49 &    96.42 &    93.56 \\
     S2 & READ\_CONTACTS                                   & -10.24 &     7.25 &    23.75 \\
     S7 & Ljava/io/IOException;-\ensuremath{>}printStackTrace                            &   7.90 &    49.82 &    66.85 \\
     S5 & android/telephony/SmsManager;-\ensuremath{>}sendTextMessage                           &   7.75 &     1.77 &    34.73 \\
     S7 & android/telephony/SmsManager;-\ensuremath{>}sendTextMessage &   7.65 &     1.77 &    34.73 \\
     S6 & INTERNET                                                          &  -4.43 &    77.74 &    85.43 \\
     S8 & ajax.googleapis.com                                                           &  -2.88 &     0.76 &     0.54 \\
     S5 & android/telephony/SmsManager;-\ensuremath{>}getDefault                                &   1.77 &     2.01 &    37.63 \\
     S7 & android/telephony/SmsManager;-\ensuremath{>}getDefault                         &   1.66 &     2.01 &    37.63 \\
\bottomrule
\end{tabular}
\end{adjustbox}
\hspace{0.05em}
\begin{adjustbox}{width=0.32\textwidth}
\begin{tabular}{cp{6cm}rrr}
\toprule
   \textbf{Set} & \textbf{Feature Name} & \multicolumn{1}{c}{$\vct r$ (\%)} & \multicolumn{1}{c}{$p_B$ (\%)} & \multicolumn{1}{c}{$p_M$ (\%)} \\
\midrule
     S2 & SEND\_SMS                           &  14.04 &     3.19 &    53.89 \\
     S4 & LAUNCHER                       & -13.65 &    96.42 &    93.56 \\
     S4 & SMS\_RECEIVED                &   8.39 &     1.09 &    20.08 \\
     S2 & RECEIVE\_SMS                        &  -8.02 &     2.43 &    38.36 \\
     S7 & android/net/Uri;-\ensuremath{>}withAppendedPath                   &  -6.96 &     9.24 &    16.96 \\
     S5 & LocationManager;-\ensuremath{>}getLastKnownLocation &  -6.45 &    27.09 &    31.65 \\
     S7 & Lorg/apache/http/client/methods/HttpPost             &   4.80 &    29.95 &    51.89 \\
     S7 & android/net/Uri;-\ensuremath{>}encode                             &  -4.64 &     9.52 &     8.17 \\
     S7 & getPackageInfo                                       &  -3.74 &    53.80 &    49.50 \\
     S2 & READ\_CONTACTS                      &  -3.57 &     7.25 &    23.75 \\
\bottomrule
\end{tabular}
\end{adjustbox}
\hspace{0.05em}
\begin{adjustbox}{width=0.32\textwidth}
\begin{tabular}{cp{6cm}rrr}
\toprule
   \textbf{Set} & \textbf{Feature Name} & \multicolumn{1}{c}{$\vct r$ (\%)} & \multicolumn{1}{c}{$p_B$ (\%)} & \multicolumn{1}{c}{$p_M$ (\%)} \\
\midrule
     S4 & LAUNCHER                            & -2.75 &    96.42 &    93.56 \\
     S2 & INSTALL\_SHORTCUT (launcher) &  2.19 &     1.51 &    26.37 \\
     S2 & ACCESS\_WIFI\_STATE                       &  1.81 &    10.59 &    43.10 \\
     S7 & TelephonyManager;-\ensuremath{>}getNetworkOperator   &  1.74 &     6.01 &    46.57 \\
     S5 & Contacts\$People;-\ensuremath{>}createPersonInMyContactsGroup & -1.64 &     3.53 &     0.89 \\
     S6 & READ\_CONTACTS                                             & -1.55 &    12.89 &     7.79 \\
     S4 & BOOT\_COMPLETED                        &  1.51 &     6.73 &    66.08 \\
     S2 & WRITE\_SETTINGS                          & -1.49 &     3.67 &    12.34 \\
     S5 & LocationManager;-\ensuremath{>}isProviderEnabled             & -1.48 &    12.53 &    17.12 \\
     S7 & android/net/Uri;-\ensuremath{>}encode                                  & -1.44 &     9.52 &     8.17 \\
\bottomrule
\end{tabular}
\end{adjustbox}
\vspace{-1.5em}
\end{table*}

\vspace{-0.2em}
\subsection{Local Explanations} Table~\ref{tab:local-ranks} reports the top-10 influential features, sorted by their (absolute) \emph{relevance} values, for three distinct samples classified by the linear \SVM and the \RF classifier, along with their probability of being present in each class.
Notably, relevant features can also be \emph{rare}. This means that a feature is deemed relevant even if it characterizes well only a small subset of samples in a given class (\eg, a malware family).

\myparagraph{Case 1} The first example is a benign application misclassified by the \SVM with a score of $-0.17$, and correctly classified by the \RF (probability $0.77\%$ and surrogate score of $+0.10$). 
By observing the features through their relevance scores, it is evident that the \RF is able to correctly classify this sample as benign as several features are assigned a negative relevance score, while almost all of them are considered as malicious (positive score) by the \SVM. In both cases the use of SMS messages for communication is retained suspicious; however, for the \RF this is not a sufficient evidence of maliciousness.

\myparagraph{Case 2} The second example is a malware sample of the \texttt{SmsWatcher} family, which is correctly classified by the \SVM (score $+0.99$), but not by the \RF model (probability $0.3\%$ and surrogate score of $-1.43$), for a reason similar to the previous case: permissions ($S_2$) and API calls ($S_7$) related to SMS usage are not a sufficient evidence of maliciousness for the \RF. Indeed, this classifier does not even identify as suspicious the application components ($S_3$) related to SMS usage, which instead constitute a \emph{signature} for this malware family, as correctly learned by the linear \SVM model.

\myparagraph{Case 3} The last case is a malware sample of the \texttt{Plankton} family, correctly classified by both models (\SVM score $+2.75$; \RF probability $0.9\%$ and surrogate score $+1.32$), as they correctly identified the behavioral patterns of this family associated to HTTP communication and actions.

\vspace{-0.2em}
\subsection{Global Explanations}

\begin{table}[t]
\centering
\caption{Top 15 malware families in the test set.}\label{tab:mal-fams}
\begin{adjustbox}{width=0.475\textwidth}
\begin{tabular}{p{1.5cm}r|p{1.5cm}r|p{1.5cm}r|p{1.5cm}r|p{1.5cm}r}
\toprule 
\textbf{Family} & \multicolumn{1}{c}{\textbf{\#}} &  \textbf{Family} & \multicolumn{1}{c}{\textbf{\#}} & 
\textbf{Family} & \multicolumn{1}{c}{\textbf{\#}} &  \textbf{Family} & \multicolumn{1}{c}{\textbf{\#}} &  \textbf{Family} & \multicolumn{1}{c}{\textbf{\#}} \\ 
\midrule 
FakeInstaller & 901 & Opfake & 591 & Iconosys & 149 & Adrd & 88 & LinuxLotoor & 69 \\ 
DroidKungFu & 640 & GingerMaster & 332 & Kmin & 144 & Geinimi & 88 & MobileTx & 68 \\ 
Plankton & 609 & BaseBridge & 318 & FakeDoc & 128 & DroidDream & 81 & GoldDream & 67 \\
\bottomrule 
\end{tabular}
\end{adjustbox}
\end{table}

We performed a global analysis of the models learned by each algorithm by averaging the local relevance vectors $\vct r$ over different classes of samples: benign, malware, and the top-$15$ malware families with the largest number of samples in the \DDrebin data (Table \ref{tab:mal-fams}). This gives us a global (mean) relevance vector $\bar{\vct r}$ for each class.
Then, for each class of samples, we report a \emph{compact} and a \emph{fine-grained} analysis of the \emph{global} feature-relevance values $\bar{\vct r}$. 
In the \emph{compact} analysis, we further average the global relevance $\bar{\vct r}$ over each \emph{feature set} $S_1, \ldots, S_8$ (Table~\ref{tab:feature_sets}).
In the \emph{fine-grained} analysis, we simply report the global relevance score $\bar{\vct r}$ for the top $44$ features (selected by aggregating the top 5 features with the highest average relevance score for each class of samples).

\begin{figure*}[t]
\centering
\includegraphics[width=\textwidth]{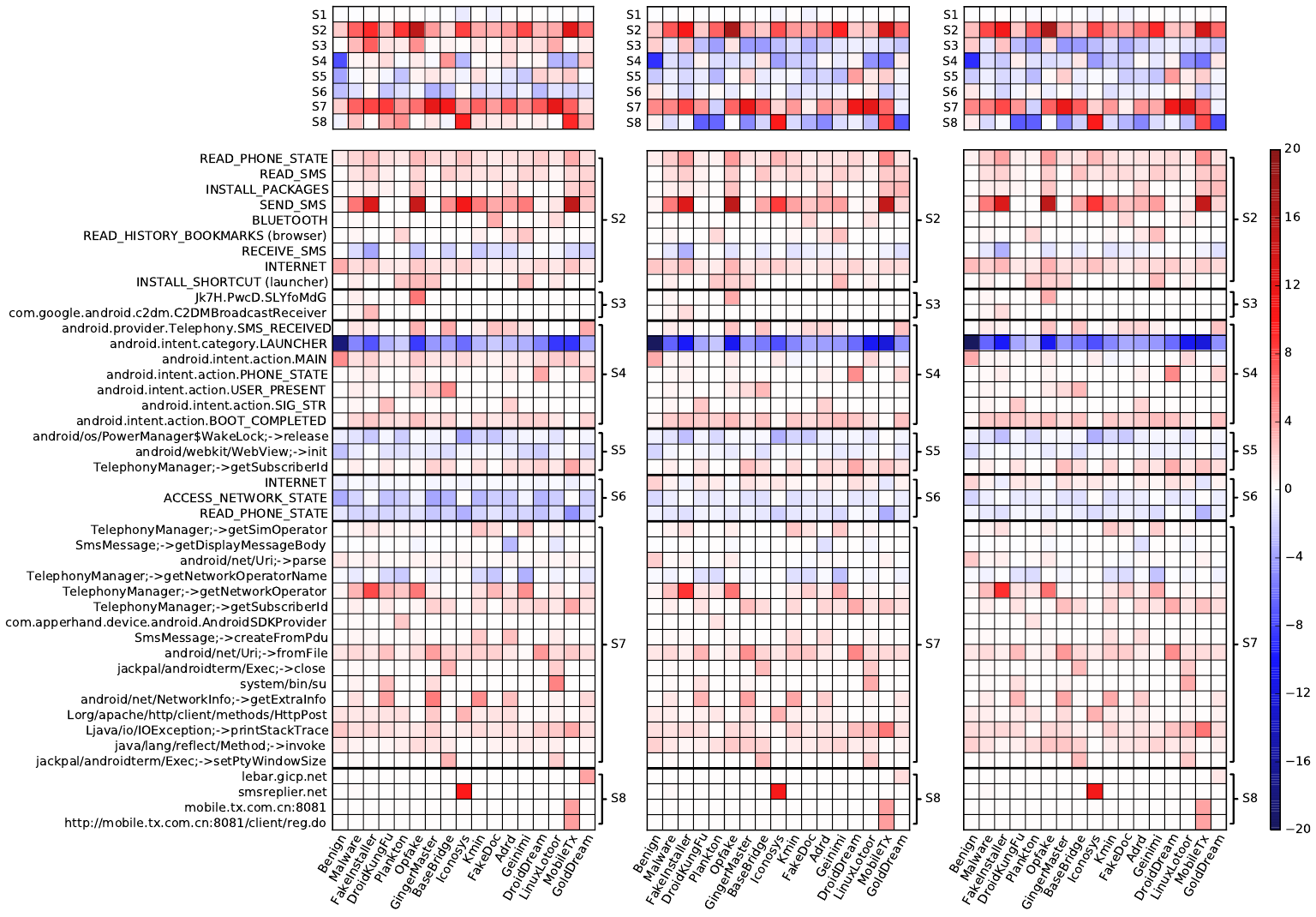}
\caption{Mean relevance scores computed w.r.t. benign, malware and the top-15 malware families (Tab.~\ref{tab:mal-fams}) for \SVM (\emph{left}), \SVMRBF (\emph{middle}) and \RF (\emph{right}). The compact representation (\emph{top}) reports the mean relevances for the feature sets $S_1, \ldots, S_8$ (Tab.~\ref{tab:feature_sets}). The fine-grained representation (\emph{bottom}) reports the mean relevances for the top $44$ features with the highest average value per family. Positive (negative) relevances denote malicious (benign) behavior.} \label{fig:rel-matrix}
\vspace{-1em}
\end{figure*}

The results are shown in Fig.~\ref{fig:rel-matrix}. The compact analysis highlights the importance of permissions ($S_2$) and suspicious API calls ($S_7$ group) in identifying malware. This is reasonable, as the majority of malware samples require permissions to perform specific actions, like stealing contacts and opening SMS and other side communication channels.
 The fine-grained analysis provides a more detailed characterization of the aforementioned behavior, highlighting how each classifier learns a specific \emph{behavioral signature} for each class of samples. In particular, malware families are characterized by their communication channels (\eg, SMS and HTTP), by the amount of stolen information and accessed resources, and by specific application components or URLs ($S_3$ and $S_8$).  

Finally, note that all classifiers tend to assign high relevance to a very small set of features in each decision, both at a local and at a global scale. Given that manipulating the content of Android malware can be relatively easy, especially due to the possibility of injecting dead code, this behavior highlights the potential \emph{vulnerability} of such classifiers. 
In fact, if the decisions of a classifier rely on few features, it is intuitive that detection can be easily evaded by manipulating only few of them, as also confirmed in previous work~\cite{demontis17-tdsc,calleja18}.
Conversely, if a model distributes relevance more evenly among features, evasion may be more difficult (\ie, require manipulating a higher number of features, which may not be always feasible).
More robust learning algorithms for these tasks have been proposed based exactly on this rationale, which has also a theoretically-sound interpretation~\cite{demontis17-tdsc}.

Another interesting point regards the \emph{transferability} of evasion attacks across different models, \ie, the fact that an attack crafted against a specific classifier may still be successful with high probability against a different one.
From our analysis, it is clear that in this case this property depends more on the available training data rather than on the specific learning algorithm: the three considered classifiers learn very similar patterns of feature relevances, as clearly highlighted in Fig.~\ref{fig:rel-matrix}, which simply means that they can be evaded with very similar modifications to the input sample.

\section{Contributions, Limitations and Future Work} \label{sect:conclusions}
In this paper, we provided a general approach to achieve explainable malware detection on Android, applicable to any black-box machine-learning model. Our explainable approach can help analysts to understand possible vulnerabilities of learning algorithms to well-crafted evasion attacks along with their transferability properties, besides providing a local and global understanding of how a machine-learning model makes its decisions.
We plan to analyze also different strategies to provide global explanations. In fact, averaging can potentially soften the contribution of features that are highly relevant only for few samples. Another interesting issue is how to choose the surrogate model to provide explanations for non-differentiable models. Some theoretical results show that, under certain assumptions, some learning algorithms can provide similar decision functions; \eg, nonlinear SVMs may reliably approximate random forests~\cite{breiman00}. Nevertheless, it is still required to investigate how different surrogate models impact the explanations provided by our approach.
These are all relevant issues towards the development of interpretable models, as required by the novel European General Data Protection Regulation (GDPR)~\cite{goodman16-gdpr}. The \emph{right of explanation} stated by GDPR imposes to develop models that are transparent with respect to their decisions. We believe that this work is a first step towards this direction. 

\vspace{-0.2em}
\section*{Acknowledgments}
This  work  was  partly  supported  by  the  EU  H2020 project  ALOHA,  under  the  European  Union's  Horizon  2020 research and innovation programme (grant no. 780788), and by the \emph{PISDAS} project, funded
by the Sardinian Regional Administration (CUP E27H14003150007).

% that's all folks

\vspace{-0.2em}

\begin{thebibliography}{10}
\bibitem{rieck14-drebin}
D.~Arp, M.~Spreitzenbarth, M.~H\"ubner, H.~Gascon, and K.~Rieck.
\newblock Drebin: Efficient and explainable detection of android malware in
  your pocket.
\newblock In {\em Proc. 21st NDSS}. The Internet Society, 2014.

\bibitem{backes17}
M.~Backes and M.~Nauman.
\newblock {LUNA:} quantifying and leveraging uncertainty in android malware
  analysis through Bayesian machine learning.
\newblock In {\em EuroS{\&}P}, pp. 204--217. {IEEE}, 2017.

\bibitem{baehrens10-jmlr}
D.~Baehrens, T.~Schroeter, S.~Harmeling, M.~Kawanabe, K.~Hansen, and K.-R.
  M\"{u}ller.
\newblock How to explain individual classification decisions.
\newblock {\em J. Mach. Learn. Res.}, 11:1803--1831, 2010.

\bibitem{biggio13-ecml}
B.~Biggio, I.~Corona, D.~Maiorca, B.~Nelson, N.~\v{S}rndi\'{c}, P.~Laskov,
  G.~Giacinto, and F.~Roli.
\newblock Evasion attacks against machine learning at test time.
\newblock In {\em ECML}, vol. 8190, {\em LNCS}, pp. 387--402. Springer, 2013.

\bibitem{biggio18}
B.~{Biggio} and F.~{Roli}.
\newblock Wild patterns: Ten years after the rise of adversarial machine
  learning.
\newblock {\em ArXiv}, 2018.

\bibitem{calleja18}
A.~Calleja, A.~Martin, H.~D. Menendez, J.~Tapiador, and D.~Clark.
\newblock Picking on the family: Disrupting android malware triage by forcing
  misclassification.
\newblock {\em Expert Systems with Applications}, 95:113 -- 126, 2018.

\bibitem{chen16-asiaccs}
S.~Chen, M.~Xue, Z.~Tang, L.~Xu, and H.~Zhu.
\newblock Stormdroid: A streaminglized machine learning-based system for
  detecting Android malware.
\newblock In {\em ASIA CCS}, pp. 377--388, 2016. ACM.

\bibitem{demontis17-tdsc}
A.~Demontis, M.~Melis, B.~Biggio, D.~Maiorca, D.~Arp, K.~Rieck, I.~Corona,
  G.~Giacinto, and F.~Roli.
\newblock Yes, machine learning can be more secure! A case study on Android
  malware detection.
\newblock {\em IEEE Trans. Dependable and Secure Computing}, In press.

\bibitem{kim17-arxiv}
F.~{Doshi-Velez} and B.~{Kim}.
\newblock {Towards A Rigorous Science of Interpretable Machine Learning}.
\newblock {\em ArXiv}, 2017.

\bibitem{goodfellow15-iclr}
I.~J. Goodfellow, J.~Shlens, and C.~Szegedy.
\newblock Explaining and harnessing adversarial examples.
\newblock In {\em ICLR}, 2015.

\bibitem{goodman16-gdpr}
B.~{Goodman} and S.~{Flaxman}.
\newblock {European Union regulations on algorithmic decision-making and a
  ``right to explanation''}.
\newblock {\em ArXiv}, 2016.

\bibitem{koh17-icml}
P.~W. Koh and P.~Liang.
\newblock Understanding black-box predictions via influence functions.
\newblock In {\em ICML}, 2017.

\bibitem{lipton16}
Z.~C. Lipton.
\newblock The mythos of model interpretability.
\newblock In {\em ICML Workshop on Human Interpretability in Machine Learning},
  pp. 96--100, 2016.

\bibitem{ribeiro16}
M.~T. Ribeiro, S.~Singh, and C.~Guestrin.
\newblock Why should i trust you?: Explaining the predictions of any
  classifier.
\newblock In {\em KDD}, pp. 1135--1144, 2016. ACM.

\bibitem{russu16-aisec}
P.~Russu, A.~Demontis, B.~Biggio, G.~Fumera, and F.~Roli.
\newblock Secure kernel machines against evasion attacks.
\newblock In {\em AISec}, pp. 59--69, 2016. ACM.

\bibitem{sommer10}
R.~Sommer and V.~Paxson.
\newblock Outside the closed world: On using machine learning for network
  intrusion detection.
\newblock In {\em IEEE Symp. Security and
  Privacy}, pp. 305--316, 2010. IEEE CS.

\bibitem{szegedy14-iclr}
C.~Szegedy, W.~Zaremba, I.~Sutskever, J.~Bruna, D.~Erhan, I.~Goodfellow, and
  R.~Fergus.
\newblock Intriguing properties of neural networks.
\newblock In {\em ICLR}, 2014.

\bibitem{szegedy14-iclr}
C.~Szegedy, W.~Zaremba, I.~Sutskever, J.~Bruna, D.~Erhan, I.~Goodfellow, and
  R.~Fergus.
\newblock Intriguing properties of neural networks.
\newblock In {\em ICLR}, 2014.

\bibitem{breiman00}
L.~Breiman.
\newblock Some infinity theory for predictor ensembles. \newblock Technical Report 579, Statistics Dept. UCB, 2000. 

\end{thebibliography}
\end{document}